\documentclass{IEEEtran}

\usepackage{mathptmx}
\usepackage{amsmath}
\usepackage{microtype}
\usepackage{graphicx}
\usepackage{float}
\usepackage{textgreek}
\usepackage[T1]{fontenc}
\usepackage[export]{adjustbox}
\usepackage[table]{xcolor}
\usepackage{moresize}
\usepackage[flushleft]{threeparttable}
\usepackage{tabularx}
\usepackage{algorithm}
\usepackage{algorithmicx}
\usepackage[noend]{algpseudocode}
\usepackage{amsthm}
\usepackage{amssymb}
\usepackage{mathtools}

\DeclareMathAlphabet{\mathcal}{OMS}{cmsy}{m}{n} 

\algnewcommand{\LineComment}[1]{\State \textcolor{gray}{\(\triangleright\)} \textcolor{gray}{#1}}

\algnewcommand\algorithmicforeach{\textbf{for each}}
\algdef{S}[FOR]{ForEach}[1]{\algorithmicforeach\ #1\ \algorithmicdo}

\let\labelindent\undefined
\usepackage[shortlabels]{enumitem}

\usepackage{cite}
\usepackage[caption=false,font=footnotesize]{subfig}
\usepackage{stfloats}

\usepackage[hang,flushmargin]{footmisc}

\makeatletter
\newcolumntype{"}{@{\hskip\tabcolsep\vrule width 1pt\hskip\tabcolsep}}

\newcommand{\thickhline}{%
\noalign {\ifnum 0=`}\fi \hrule height 1pt
\futurelet \reserved@a \@xhline
}
\makeatother

\newcommand{\codename}{MATIC}
\newcommand{\codenamelong}{Memory Adaptive Training and In-situ Canaries}

\begin{document}

\bstctlcite{IEEEexample:BSTcontrol}
\pagenumbering{gobble}

\title{MATIC: Learning Around Errors for Efficient Low-Voltage Neural Network Accelerators}
\author{
\IEEEauthorblockN{
   Sung~Kim$^\dagger$, Patrick~Howe$^\dagger$, Thierry~Moreau$^\ddagger$,
   Armin~Alaghi$^\ddagger$, Luis~Ceze$^\ddagger$, Visvesh~Sathe$^\dagger$ \\
   {
   \small
   Department of Electrical Engineering$^\dagger$, Paul G. Allen School of Computer Science and Engineering$^\ddagger$, University of Washington, USA \\
   \{sungk9, pdh4\}@ee.washington.edu, \{moreau, armin, ceze\}@cs.washington.edu, sathe@ee.washington.edu
   }
}
}
\maketitle %

\begin{abstract}
   \boldmath
   As a result of the increasing demand for deep neural network (DNN)-based services, efforts to develop dedicated hardware accelerators for DNNs are growing rapidly.
   However, while accelerators with high performance and efficiency on convolutional deep neural networks (Conv-DNNs) have been developed, less progress has been made with regards to fully-connected DNNs (FC-DNNs).
   In this paper, we propose MATIC (Memory Adaptive Training with In-situ Canaries), a methodology that enables aggressive voltage scaling of accelerator weight memories to improve the energy-efficiency of DNN accelerators.
   To enable accurate operation with voltage overscaling, MATIC combines the characteristics of destructive SRAM reads with the error resilience of neural networks in a memory-adaptive training process.
   Furthermore, PVT-related voltage margins are eliminated using bit-cells from synaptic weights as in-situ canaries to track runtime environmental variation.
   Demonstrated on a low-power DNN accelerator that we fabricate in 65 nm CMOS, MATIC enables up to 60-80 mV of voltage overscaling (3.3$\times$ total energy reduction versus the nominal voltage), or 18.6$\times$ application error reduction.
\end{abstract}

\section{Introduction}\label{sec:intro}
\noindent DNNs have demonstrated state-of-the-art performance on a variety of signal processing tasks, and there is growing interest in DNN hardware accelerators for application domains ranging from IoT to the datacenter.
However, much of the recent success with DNN-based approaches has been attributed to the use of larger model architectures (i.e., models with more layers), and state-of-the-art model architectures can have millions or billions of trainable weights.
As a result of weight storage requirements, neural network algorithms are particularly demanding on memory systems; for instance, the authors of \cite{diannao} found that main memory accesses dominated the total power consumption for their accelerator design.
Subsequently, \cite{shidiannao} developed an accelerator that leveraged large quantities of on-chip SRAM, such that expensive off-chip DRAM accesses could be eliminated.
To a similar end, \cite{eie} used pruning and compression techniques to yield sparse weight matrices, and \cite{eyeriss} leveraged data-reuse techniques and run-length compression to reduce off-chip memory access.
Nevertheless, once off-chip memory accesses are largely eliminated, on-chip SRAM dedicated to synaptic weights can account for greater than 50\% of total system power~\cite{eie}.
The on-chip memory problem is particularly acute in DNNs with dense classifier layers, since classifier weights are typically unique and constitute greater than 90\% of total weight parameters~\cite{alexnet}.
As a result, convolutional data-reuse techniques like those proposed in~\cite{eyeriss} and \cite{chain_nn} lose effectiveness.

Voltage scaling can enable significant reduction in static and dynamic power dissipation, however read and write stability constraints have historically prevented aggressive voltage-scaling on SRAM.
While alternative bit-cell topologies can be used, they typically trade-off bit-cell stability for non-trivial overheads in terms of area, power, or speed.
Hence, designs that employ SRAM either place on-chip memories on dedicated supply-rails, allowing them operate hundreds of millivolts higher than the rest of the design, or the system shares a unified voltage domain.
In either case, significant energy savings from voltage scaling remain unrealized due to SRAM operating voltage constraints; this translates to either shorter operating lifetime for battery-powered devices, or higher operating costs.
Furthermore, to account for PVT-variation, designers either apply additional static voltage margin, or add dummy logic circuits that detect imminent failure (\textit{canaries}).

\begin{figure}[t]
\centering
\includegraphics[width=\columnwidth]{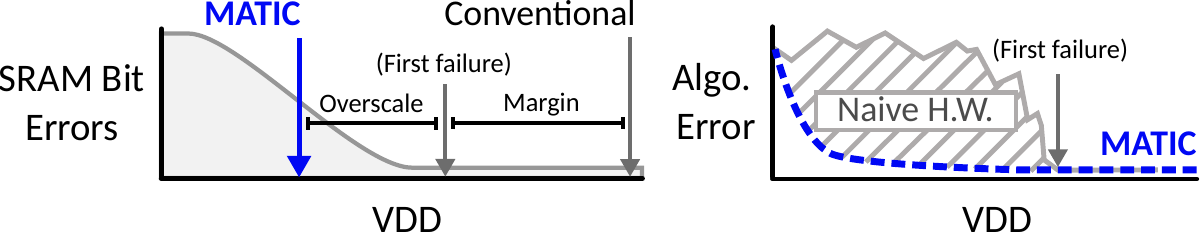}
\caption{
\codename{} enables margin reduction with in-situ canaries, and recovers from errors caused by voltage overscaling with memory-adaptive training.
}
\label{fig:snowflake5}
\end{figure}

In this paper we present \codenamelong{} (\codename{}, Figure \ref{fig:snowflake5}), a hardware/algorithm co-design methodology that enables aggressive voltage scaling of weight SRAMs with tuneable accuracy-energy tradeoffs.
To achieve this end, \codename{} jointly exploits the inherent error tolerance of DNNs with the specific characteristics of SRAM read-stability failures.
To evaluate the effectiveness of \codename{} we also design and implement SNNAC, a low-power DNN accelerator for mobile devices fabricated in 65 nm CMOS, and demonstrate state-of-the-art energy-efficiency on classification and regression tasks.

The remaining sections are organized as follows.
Section~\ref{sec:background} provides background on DNNs and failure modes in 6T SRAM.
Section~\ref{sec:method} gives the algorithmic details and design of \codename{}, and Section~\ref{sec:snnac} describes the prototype chip.
Results and comparison to prior works are discussed in Sections \ref{sec:snnac_results} and \ref{sec:related}, respectively.

\section{Background} \label{sec:background}

\noindent This section briefly reviews relevant background on DNN operation, and SRAM read-stability failure.

\subsection{Deep Neural Networks}
\noindent DNNs are a class of bio-inspired machine learning models that are represented as a directed graph of neurons~\cite{bishop_ml}.
During inference, a neuron $k$ in layer $j$ implements the composite function: $$\textstyle{z^{(j)}_k=f\left ( \sum_{i=1}^{N^{(j-1)}} w^{(j)}_{k,i}~z^{(j-1)}_i \right )},$$
where $z^{(j-1)}_i$ denotes the output from neuron $i$ in the previous layer, and $w^{(j)}_{k,i}$ denotes the weight in layer $j$ from neuron $i$ in the previous layer to neuron $k$.
$f(x)$ is a non-linear function, typically a sigmoidal function or rectified linear unit (ReLU).
Since the computation of a DNN layer can be represented as a matrix-vector dot product (with $f(x)$ computed element-wise), DNN execution is especially amenable to dataflow hardware architectures designed for linear algebra.


Training involves iteratively solving for weight parameters using a variant of gradient descent.
Given a weight $w^{(j)}_{i,k}$, its value at training iteration $n+1$ is given by: \\
$$\textstyle{w^{(j)}_{k,i}[n+1] = w^{(j)}_{k,i}[n] - \alpha \frac{\partial J}{\partial w^{(j)}_{k,i}[n]}},$$
where $\alpha$ is the step size and $J$ is a suitable loss function (e.g., cross-entropy)~\cite{bishop_ml}.
The partial derivatives of the loss function with respect to the weights are computed by propagating error backwards via partial differentiation (\textit{backprop}).
\codename{} relies on the observation that backprop makes error caused by artificial weight perturbations observable, and subject to compensation via weight adaptation.
\subsection{SRAM Read Failures}

\begin{figure}[t]
\centering
\includegraphics[width=0.7\columnwidth]{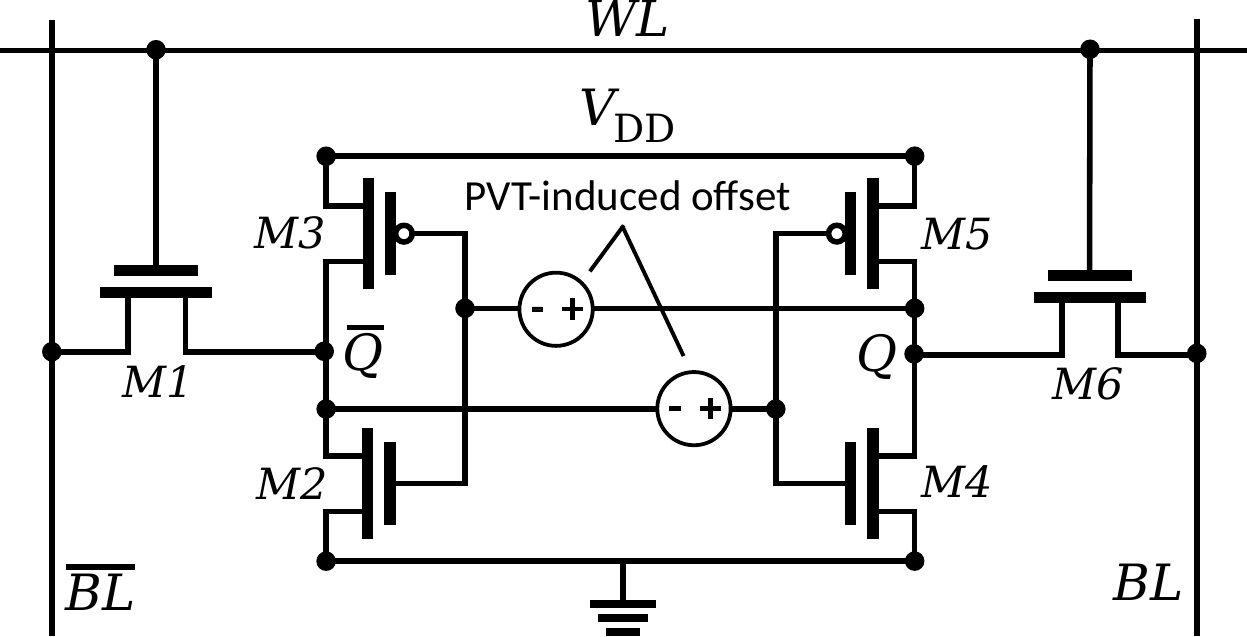}
\caption{
   An SRAM 6T bit-cell with mismatch-induced, input-referred static offsets.
   During a read, the active pull-down device (either M2 or M4) may be overcome by the current sourced from the pre-charged bit-line (via the access device) if there is insufficient static noise margin.
}
\label{fig:bitcell}
\end{figure}


\noindent Figure~\ref{fig:bitcell} shows a 6T SRAM bit-cell that is composed of two cross-coupled inverters (M2/M3 and M4/M5) and access transistors (M1 and M6).
Variation-induced mismatch between devices in the bit-cell creates an inherent, state-independent offset~\cite{guo_jssc_2009}.
This offset results in each bit-cell having a ``preferred state.''
For instance, the bit-cell depicted in Figure~\ref{fig:bitcell} favors driving $Q$ and $\overline{Q}$ to logic `1' and `0', respectively.
Due to statistical variation, larger memories are likely to see greater numbers of cells with significant offset error.

As supply voltage scales, the diminished noise margin allows the bit-cell to be flipped to its preferred state during a read~\cite{guo_jssc_2009}.
This type of read-disturbance failure, which occurs at the voltage $V_{min,read}$, is a result of insufficient pull-down strength in either M2 or M4 (whichever side of the cell stores 0) relative to the pass-gate and pre-charged bit-line.

Once flipped, the bit-cell retains state in subsequent repeated reads, favouring its (now incorrect) bit value due to the persistence of the built-in offset.
Consequently, the occurrence of bit-cell read-stability failure at low supply voltages is random in space, but essentially provides {\em stable} read outputs consistent with its preferred state~\cite{guo_jssc_2009}.
Notably, the read failures described above are in distinct from bit-line access-time failures, which can be corrected with ample timing margin.

\section{Voltage Scaling for DNN Accelerators} \label{sec:method}

\begin{figure}[t]
\centering
\includegraphics[width=\columnwidth]{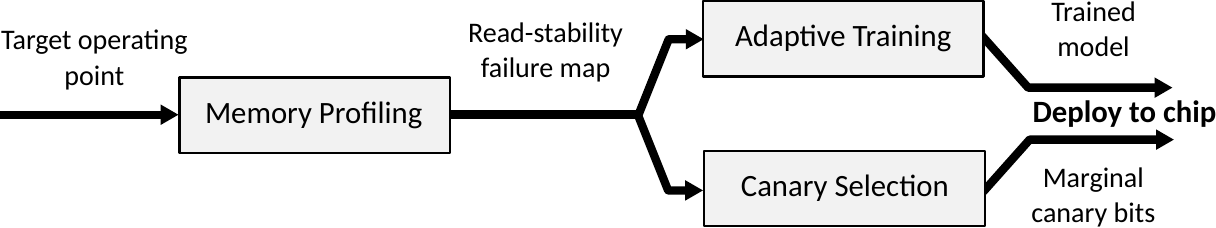}
\caption{
Overview of the \codename{} compilation and deployment flow.
}
\label{fig:snowflake4}
\end{figure}

\noindent We now present \codename{}, a voltage scaling methodology that leverages memory-adaptive training (MAT) to operate SRAMs past their point of failure, and in-situ synaptic canaries to remove static voltage margins for accurate operation across PVT variation.
In conjunction, the two techniques enable system-wide voltage scaling for energy-efficient operation, with tuneable accuracy-energy tradeoffs.
Figure \ref{fig:snowflake4} shows an overview of the processing and deployment flow, which is detailed below.

\subsection{SRAM Profiling}
\noindent SRAM read failures are profiled post-silicon to be used later during the memory-adaptive training process and in-situ canary selection.
The SRAM profiling procedure takes place once at compile time, and consists of a read-after-write and read-after-read operation on each SRAM address, at the target DNN accuracy level (bit-error proportion).
The word address, bit index, and error polarity of each bit-cell failure are then collected to form a complete failure map for each voltage-scalable weight memory in the hardware design.
The entire profiling process and failure-map generation is automated with a host PC that controls on-chip debug software and external digitally-programmable voltage regulators.

\subsection{Memory-Adaptive Training}

\begin{figure}[t]
\centering
\includegraphics[width=\columnwidth]{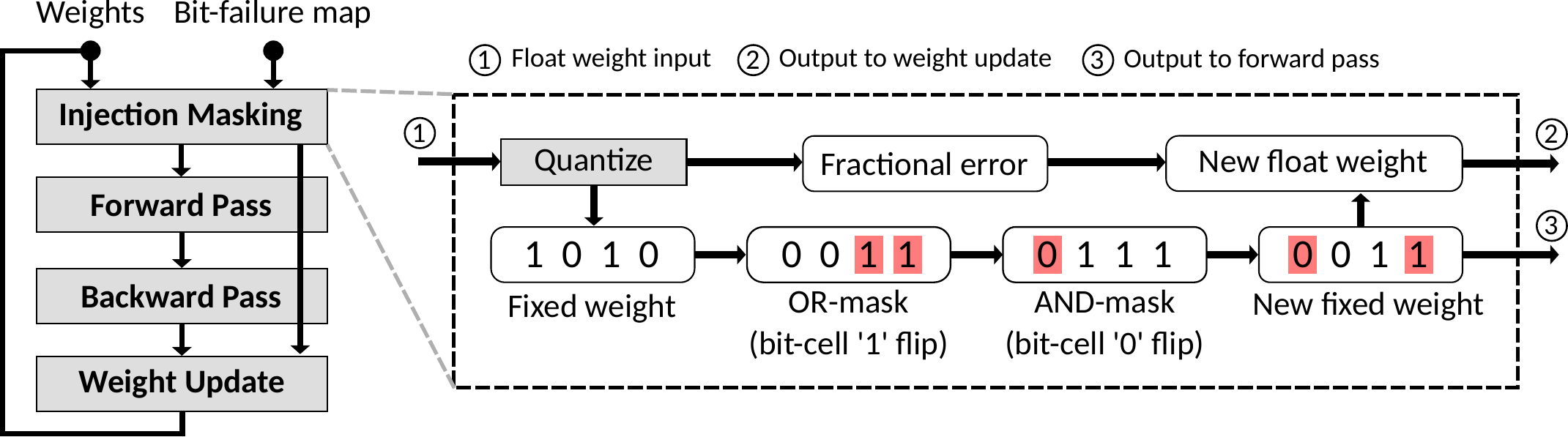}
\caption{
The modified DNN training algorithm and injection masking process.
\texttt{OR} and \texttt{AND} bit masks annotate information on SRAM bit-errors that occur due to voltage overscaling,
and quantization convergence issues are avoided by maintaining both float and fixed-point weights.
}
\label{fig:inj_mask}
\end{figure}

\begin{figure}[t]
\centering
\includegraphics[width=\columnwidth]{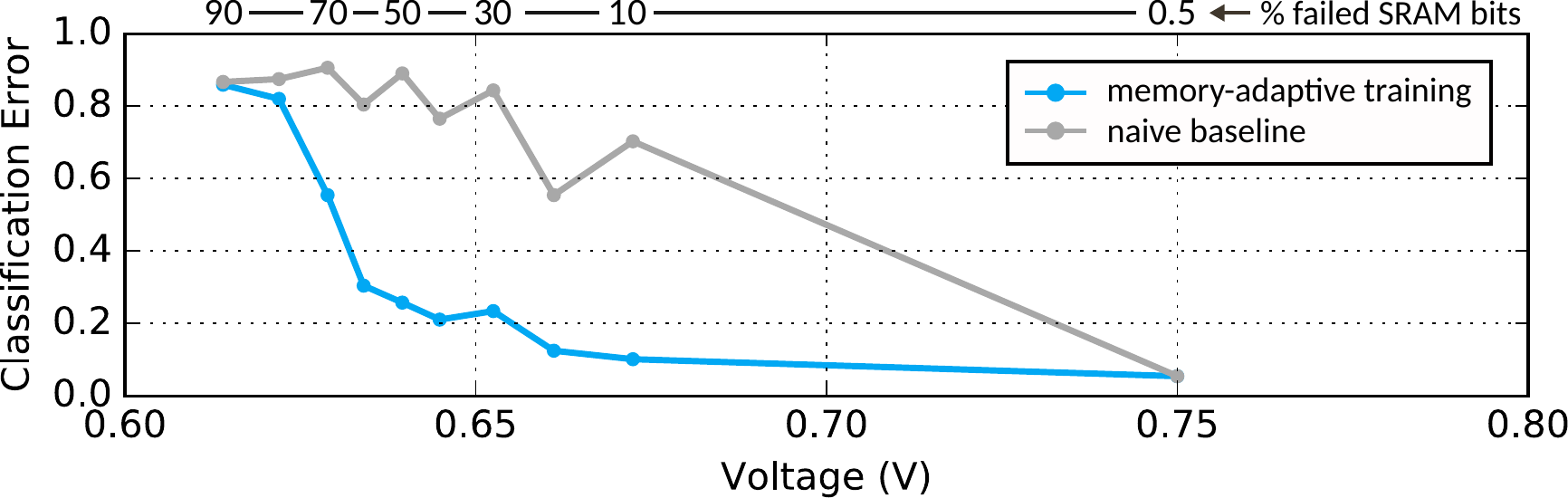}
\caption{
Simulated performance of memory-adaptive training on MNIST.
}
\label{fig:zfrac}
\end{figure}

\noindent MAT augments the vanilla backprop algorithm by injecting profiled SRAM bit-errors into the training process, enabling the neural network to compensate via adaptation.
As described in Section \ref{sec:background}, random mismatch results in bit-cells that are statically biased towards a particular storage state.
If a bit-cell stores the complement of its ``preferred'' state, performing a read at a sufficiently low voltage flips the cell and subsequent reads will be incorrect, but stable.
MAT leverages this stability during training with an \textit{injection masking} process (Figure~\ref{fig:inj_mask}).
The injection mask applies bit-masks corresponding to profiled bit-errors to each DNN weight before the forward training pass.
As a result, the network error propagated in the backward pass reflects the impact of the bit-errors, leading to compensatory weight updates in the entire network.
Since the injection masking process operates on weights that correspond to real hardware, weights are necessarily quantized during training to match the SRAM word length.
However, previous work has shown that unmitigated quantization during training can lead to significant accuracy degradation\cite{stochastic_rounding}.
MAT counteracts the effects of quantization during training by preserving the fractional quantization error, in effect performing floating point training to enable gradual weight-updates that occur over multiple backprop iterations.
The augmented weight update rule for MAT is given by

$$\textstyle{w^{(j)}_{k,i}[n+1] = m^{(j)}_{k,i}[n] - \alpha \frac{\partial J}{\partial m^{(j)}_{k,i}[n] } + \epsilon_q },~\text{and}$$
$$\textstyle{m^{(j)}_{k,i}[n]  = B_{or} B_{and} Q\left(w^{(j)}_{k,i}[n]\right)}, $$

\noindent in which $m^{(j)}_{k,i}$ is the quantized weight that corresponds to $w^{(j)}_{k,i}$, $B_{or}$ and $B_{and}$ are the bit-masks corresponding to bit-cell faults in the physical SRAM word, $Q$ is the quantization function, and $\epsilon_q$ is the fractional quantization error.

To evaluate the feasibility of memory-adaptive training, we first examine the memory-adaptive training flow with simulated SRAM failure patterns.
We implement the training modifications described above in the open-source FANN\cite{fann} and Caffe\cite{caffe} frameworks.
A proportion of randomly selected weight bits are statically flipped at each voltage, where the proportion of faulty bits is determined from SPICE Monte Carlo simulations of a 6T bit-cell.
Figure~\ref{fig:zfrac} shows that a significant fraction of bit errors can be tolerated, and that \codename{} provides a reasonably smooth energy-error tradeoff curve.

\begin{figure}[t]
\centering
\includegraphics[width=\columnwidth]{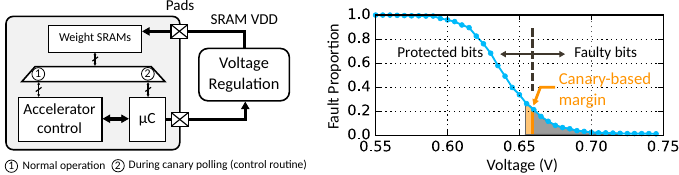}
\caption{
Overview of the runtime SRAM-voltage control scheme, using bit cells from synaptic weights as in-situ canaries.
Between inferences, the integrated \textmu C polls canary bits to determine if voltage modifications should be applied.
}
\label{fig:canary}
\end{figure}

\begin{algorithm}[t]
   \caption{In-Situ Canary-based Voltage Control}
   \label{alg:canary}
   \small
   \begin{algorithmic}
      \State $\mathcal C \coloneqq $ Set of marginal canary bits
      \State $v_0 \coloneqq $ Safe initial voltage
      \LineComment{... controller wakes from sleep}
      \State $SetSRAMVoltage(v_0)$
      \Repeat
      \State $SetSRAMVoltage(v - \Delta v)$
      \State $any\_failed \gets CheckStates(\mathcal C)$
      \If {$any\_failed$}
         \State $SetSRAMVoltage(v + \Delta v)$
         \State $RestoreStates(\mathcal C)$
      \Else
         \State $v \gets v - \Delta v$
      \EndIf
      \Until {$any\_failed$}
      \LineComment{... controller returns to sleep}
   \end{algorithmic}
\end{algorithm}

\subsection{In-Situ Synaptic Canaries}

\noindent The in-situ canary circuits are bit-cells selected directly from weight SRAMs that facilitate SRAM supply-voltage control at runtime~(Figure~\ref{fig:canary}).
Traditional canary circuits replicate critical circuits to detect imminent failure, but require added margin and are vulnerable to PVT-induced mismatch.
Instead, \codename{} uses weight bit-cells directly as in-situ canary circuits, leveraging a select number of bit-cells that are on the margin of read-failure.
This maintains a target bit-cell fault \textit{pattern}, and in turn maintains the level of classification accuracy.
In contrast to a static voltage margin, the in-situ canary-based margin provides reliability tailored to the specific failure patterns of the test chip.
The in-situ canary technique relies on two key observations:
\begin{itemize}[leftmargin=*]
\item[1.] Since the most marginal, failure prone bit-cells are chosen as canaries, canaries fail before other performance-critical bit-cells and protect their storage states.
\item[2.] Neural networks are inherently robust to a limited number of \textit{uncompensated} errors\cite{temam_defect_tol}. As a result, network accuracy is not dependent on the failure states of canary bit-cells, and the actual operating voltage can be brought directly to the $V_{min,read}$ boundary of the canaries.
\end{itemize}

At runtime, in-situ canary bits are polled by a runtime controller to determine whether supply voltage modifications should be applied.
While we use an integrated microcontroller in the test chip described in Section \ref{sec:snnac_results}, the runtime controller can be implemented with faster or more efficient circuits.
For canary selection and voltage adjustment, we conservatively select eight distributed, marginal canary bit-cells from each weight-storage SRAM.
The in-situ canary-based voltage control routine is summarized in Algorithm \ref{alg:canary}.

\section{DNN Accelerator Architecture}\label{sec:snnac}

\begin{figure}[t]
   \centering
   \begin{minipage}{0.34\linewidth}
      \centering
      \subfloat[]{\includegraphics[width=0.9\columnwidth]{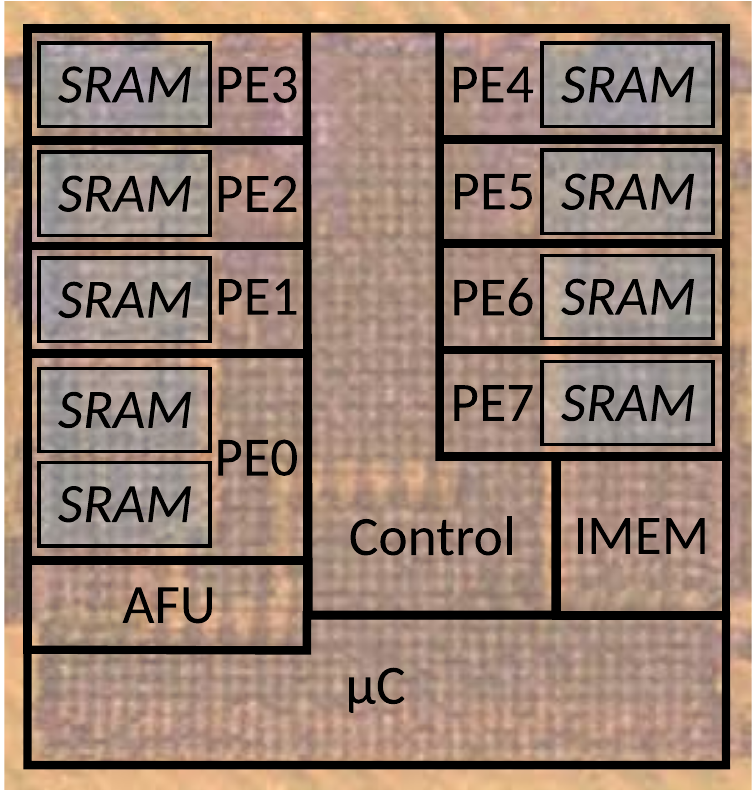}}
   \end{minipage}
   \hspace{0.02\linewidth}%
   \begin{minipage}{0.57\linewidth}
      \centering
      \subfloat[]{
         \scriptsize
         \begin{adjustbox}{max width=\textwidth}
         {\renewcommand{\arraystretch}{1.4}%
         \begin{tabular}{c"c}
            \hline
            Technology & TSMC GP 65 nm  \\
            \hline
            Core Area & 1.15$\times$1.2 mm \\
            \hline
            SRAM & 9 KB \\
            \hline
            Voltage & 0.9 V \\
            \hline
            Frequency & 250 MHz \\
            \hline
            Power & 16.8 mW \\
            \hline
            Energy & 67.1 pJ/cycle \\
            \hline
         \end{tabular}
         }
         \end{adjustbox}
      }
   \end{minipage}%
   \caption{%
   \textbf{(a)} Microphoto of a fabricated SNNAC test chip, and \textbf{(b)} nominal performance characteristics.
   }
\label{fig:summary_and_die}
\end{figure}

\begin{figure}[t]
\centering
\begin{center}
\includegraphics[width=\columnwidth]{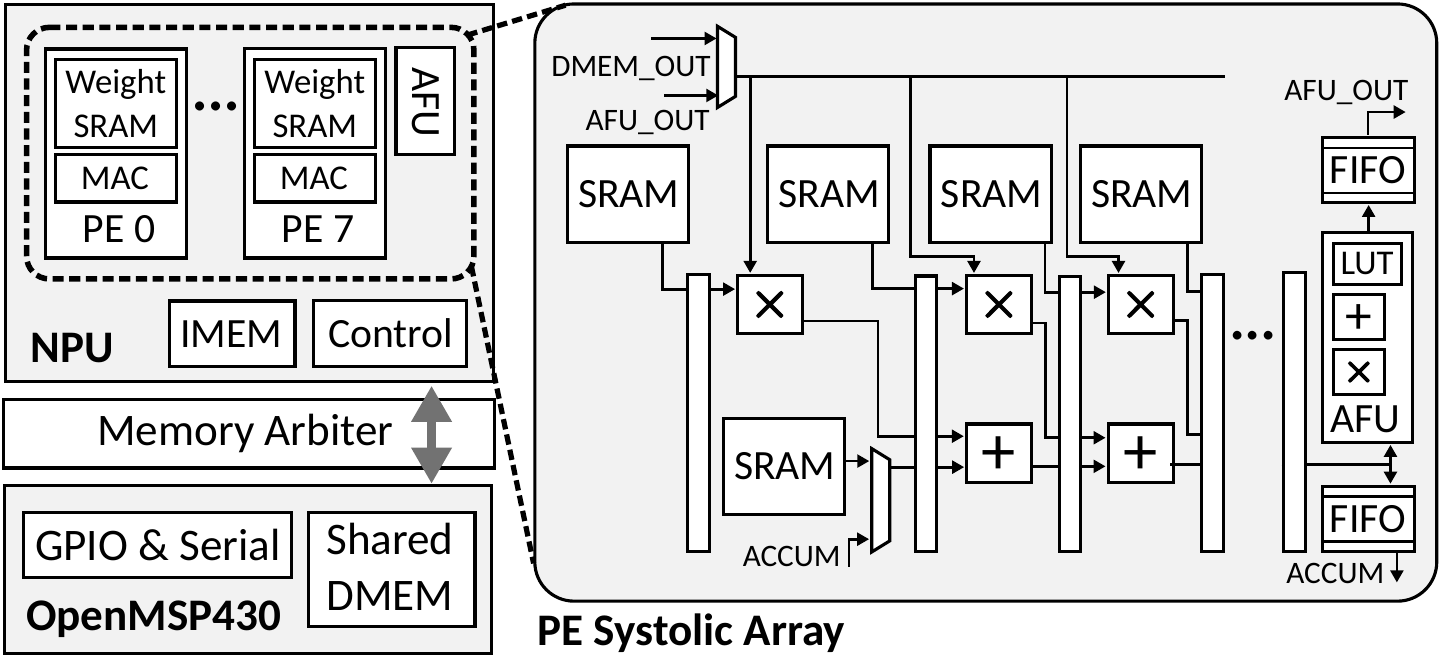}
\end{center}
\vspace{-0.2cm}
\caption{
Architecture of the SNNAC DNN accelerator.
}
\label{fig:hw_blocks}
\end{figure}

\noindent To demonstrate the effectiveness of \codename{} on real hardware, we implement SNNAC (Systolic Neural Network AsiC) in 65 nm CMOS technology (Figure \ref{fig:summary_and_die}).
The SNNAC architecture~(Figure \ref{fig:hw_blocks}) is based on the open-source systolic dataflow design from SNNAP~\cite{snnap}, optimized for integration with a light-weight SoC.

The SNNAC core consists of a fully-programmable Neural Processing Unit (NPU) that contains eight multiply-accumulate (MAC)-based Processing Elements (PEs).
The PEs are arranged in a 1D systolic ring that maintains high compute utilization during inner-product operations.
Energy-efficient arithmetic in the PEs is achieved with 8-22 bit fixed-point operands, and each PE includes a dedicated voltage-scalable SRAM bank to enable local storage of synaptic weights.
The systolic ring is attached to an activation function unit (AFU), which minimizes energy and area footprint with piecewise-linear approximation of activation functions (e.g., sigmoid or ReLU).

The operation of the PEs is coordinated by a lightweight control core that executes statically compiled microcode.
To achieve programmability and support for a wide range of layer configurations, the computation of wide DNN layers is time-multiplexed onto the PEs in the systolic ring.
When the layer width exceeds the number of physical PEs, PE results are buffered to an accumulator that computes the sum of all atomic MAC operations in the layer.
SNNAC also includes a sleep-enabled OpenMSP430-based microcontroller (\textmu C) \cite{openmsp} to handle runtime control, debugging functions, and off-chip communication with a UART serial interface.
To minimize data movement, NPU input and output data buffers are memory-mapped directly to the \textmu C data address space.

\begin{figure}[t]
\centering
\vspace{-0.2in}
\subfloat[]{\includegraphics[width=0.36\columnwidth]{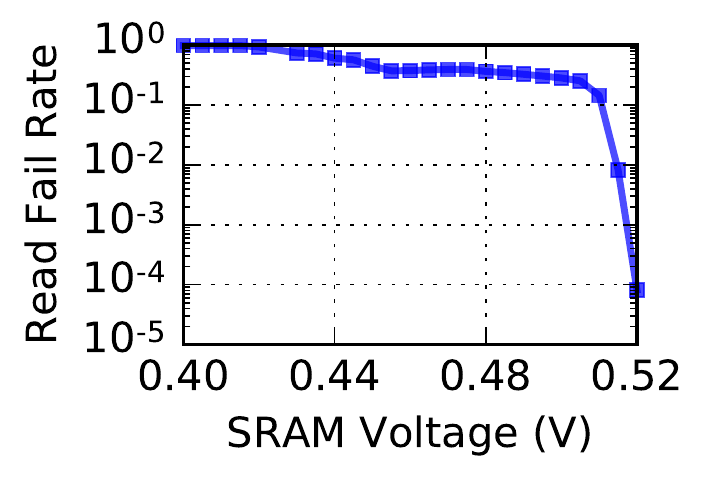}}
\subfloat[]{\includegraphics[width=0.64\columnwidth]{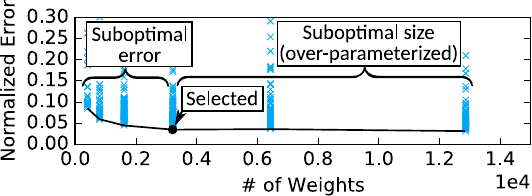}}
   \caption{\textbf{(a)} Measured SRAM read-failure rate at 25$^\circ$C. \textbf{(b)} Topology selection to avoid biased overparameterization - each point is a unique DNN topology.}
\label{fig:read_fail_and_knee}
\end{figure}

\begin{figure}[t]
\centering
\begin{center}
\includegraphics[width=\columnwidth]{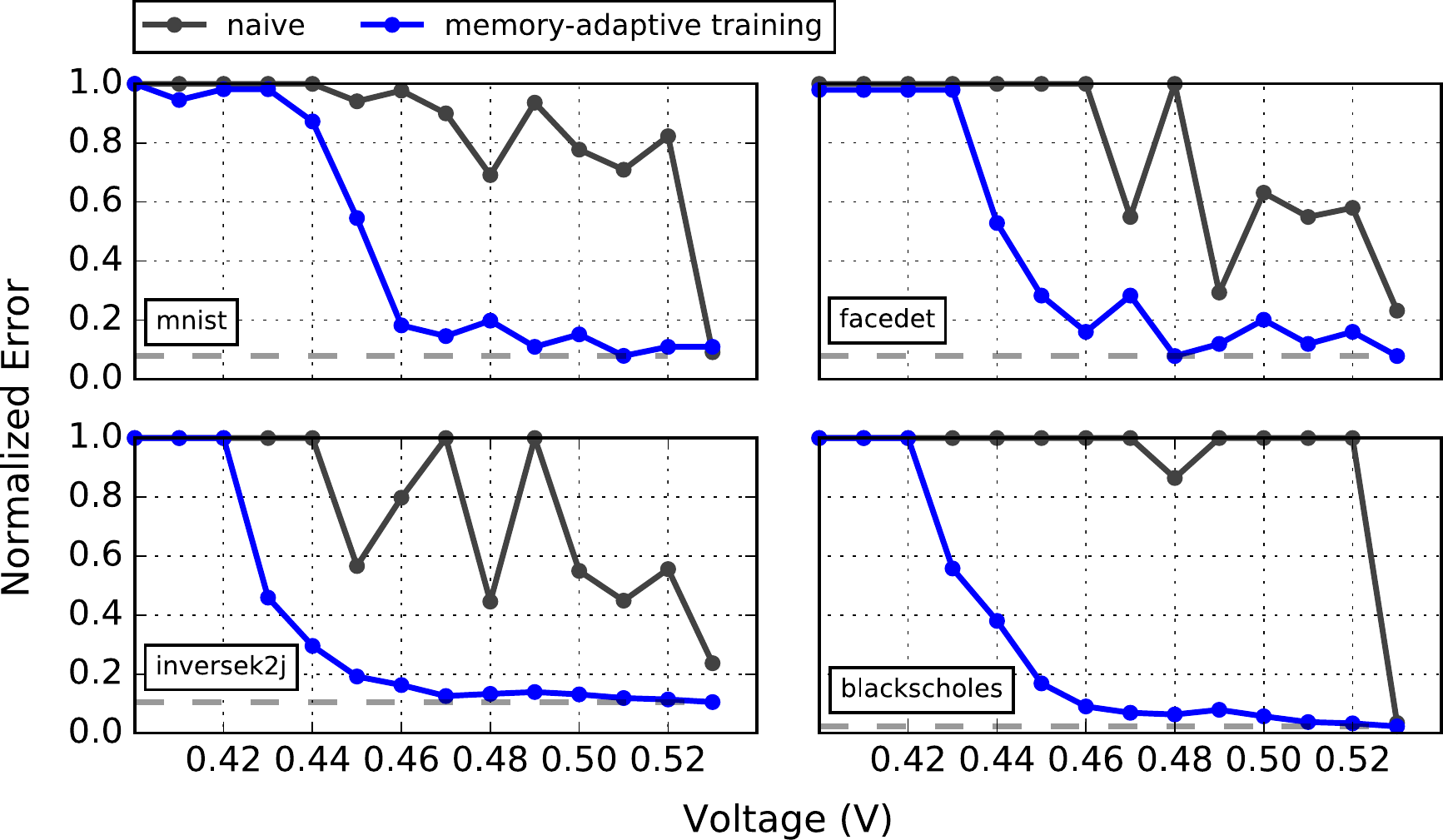}
\end{center}
\vspace{-0.2cm}
\caption{
Error performance of SNNAC, with and without MATIC deployed.
}
\label{fig:error_grid}
\end{figure}

\section{Experimental Setup and Hardware Results}\label{sec:snnac_results}

\noindent At 25$^\circ$C and 0.9 V, a nominal SNNAC implementation operates at 250 MHz and dissipates 16.8 mW, achieving a 90.6\% classification rate on MNIST handwritten character recognition~\cite{mnist}.
In addition to MNIST, we evaluate face detection on the MIT CBCL face database~\cite{mit_faces}, and two approximate computing benchmarks from~\cite{esmael_2012}.
For all of the benchmark tasks, we divide the datasets into training and test subsets with either a 7-to-1 or 10-to-1 train/test split.

We find that the compiled SRAMs (rated at 0.9 V) exhibit bit-errors starting from 0.53 V at room temperature, with all reads failing at \textasciitilde0.4 V (Figure \ref{fig:read_fail_and_knee}a).
Since the point of first failure is dictated by the tails of the $V_{min,read}$ statistical distribution, we expect voltage savings to increase in more advanced process nodes, and with larger memories.
For instance, the SRAM variability study from \cite{guo_jssc_2009} exhibits $V_{min,read}$ failures starting at \textasciitilde0.66 V with a 45 nm, 64~kb array.

\begin{table*}
\centering
   \caption{DNN benchmarks and application error measurements.}
   \label{tab:app_error}
\resizebox{42pc}{!}{%
\begin{tabular}{c"c|c|c"c"c|c"c|c"c|c"c}
   \hline
   Benchmark               & Description         & Error Metric   & DNN                     & Error@0.9 V   & Error@0.50 V  & Error@0.50 V  & Error@0.46 V  & Error@0.46 V  &   AEI   &   AEI    & Reduction \\
                           &                     &                & Topology                & (nominal)    & (naive)      & (adaptive)   & (naive)      & (adaptive)   & (naive) & (adapt.) &  \\
   \thickhline
   \textit{mnist} \cite{mnist}         &Digit recognition   & Classif. rate  & 100-32-10    &  9.4\%       &  70.7\%      & 13.0\%       & 84.0\%       & 15.6\%       & 62.5\%   & 5.0\% & \textbf{12.5}  \\
   \textit{facedet} \cite{mit_faces}   &Face detection      & Classif. rate  & 400-8-1      &  12.5\%      &  33.6\%      & 15.6\%       & 47.7\%       & 15.8\%       & 37.5\%   & 5.6\% & \textbf{6.7}  \\
   \textit{inversek2j} \cite{esmael_2012}  &Inverse kinematics  & Mean sq. error & 2-16-2       &  0.032       &  0.169       & 0.040        & 0.245        & 0.050        & 50.7\%   & 1.9\% & \textbf{26.7} \\
   \textit{bscholes} \cite{esmael_2012}    &Option pricing      & Mean sq. error & 6-16-1       &  0.021       &  0.094       & 0.023        & 0.094        & 0.026        & 65.3\%   & 2.3\% & \textbf{28.4} \\
   \hline
   Average                  & -                  & -              & -       &      - &    -    &  - &      -     &     - & -     & -     & \textbf{18.6} \\

   \hline
\end{tabular}
   }
\end{table*}

\subsection{Application Error}
\noindent Figure \ref{fig:error_grid} shows how \codename{} recovers application error after the point of first failure.
Compared to a voltage-scaled naive system (the SNNAC accelerator operating with baseline DNN models), \codename{} demonstrates much lower application error.
The baseline and memory-adaptive models use the same DNN model topologies (e.g., layer depth and width configurations), but memory-adaptive training modifications are disabled for the naive case.
To avoid unfair bias in the application error analysis, all benchmarks use \textit{compact} DNN topologies that minimize intrinsic over-parameterization (Figure \ref{fig:read_fail_and_knee}b).
Table \ref{tab:app_error} lists the benchmarks along with model descriptions, and application error for the baseline (naive) and memory-adaptive evaluations.
We list the application error at the nominal voltage (0.9 V), in addition to the energy-optimal voltage (0.5 V), and 0.46 V, which is the voltage where application error increases significantly.
Between 0.46 V and 0.53 V, the use of MATIC results in 6.7$\times$ to 28.4$\times$ application error reduction versus naive hardware.
When averaged across both voltage and all benchmarks, the average error-increase (AEI) is reduced~by~18.6$\times$.

\subsection{Energy-Efficiency}

\begin{figure}[t]
\centering
\includegraphics[width=\columnwidth]{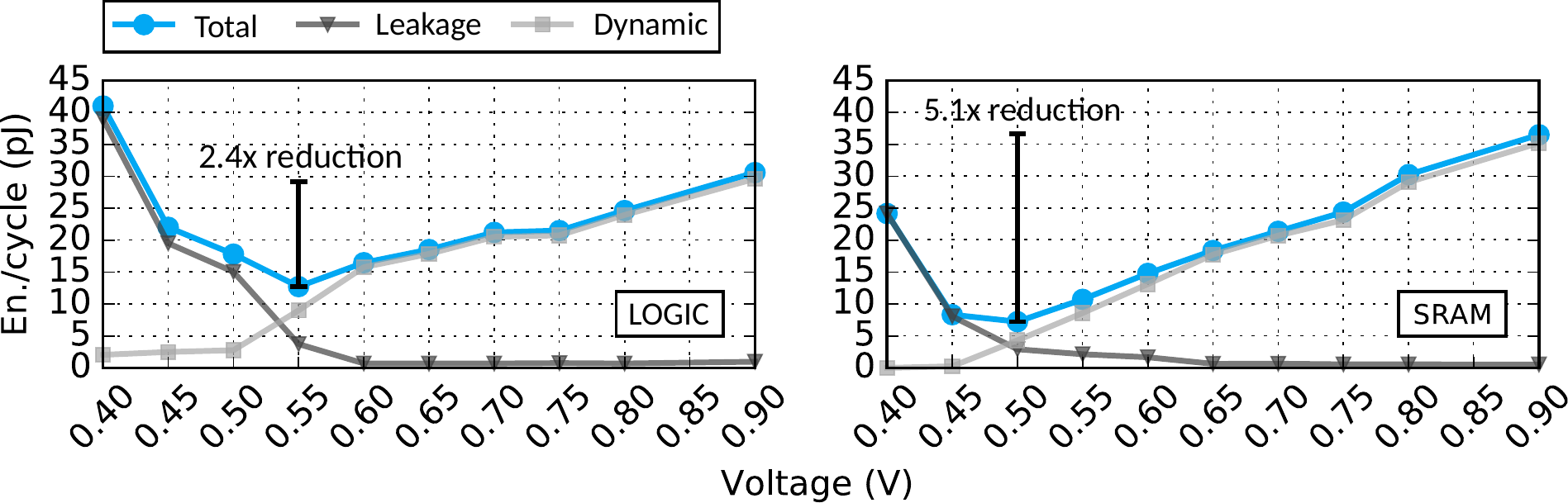}
\caption{
Energy-per-cycle measurements for SNNAC, obtained from test chip current measurements.
}
\label{fig:energy}
\end{figure}

\begin{table}[t]
\centering
\caption{
Energy-efficiency with MATIC-enabled scaling.
}
\label{tab:energy}
\resizebox{\columnwidth}{!}{%
\begin{tabular}{c"cc|cc|cc}
   \hline
   Param/Config. & {\bf \textit{HighPerf}} & Base &  {\bf \textit{EnOpt\_split}} & Base &  {\bf \textit{EnOpt\_joint}} & Base \\
   \thickhline
   Logic Voltage (V) & 0.9 & 0.9 & 0.55 & 0.55 & 0.55 & 0.9 \\
   SRAM Voltage (V) & 0.65 & 0.9 & 0.5 & 0.9 & 0.55 & 0.9 \\
   \hline
   Frequency (MHz) & 250 & - & 17.8 & - & 17.8 & - \\
   \hline
   {\bf Total Energy (pJ/cycle)} & {\bf 48.96}   & {\bf 67.08} & \underline{{\bf 19.98}}   & {\bf 49.23} & {\bf 20.60}   & {\bf 67.08} \\
   Logic       & 30.58        & 30.58     & 12.73        & 12.73     & 12.73        & 30.58 \\
   SRAM        & 18.37        & 36.50     & 7.24         & 36.50     & 7.86         & 36.50 \\
   \hline
   \textbf{Energy Reduction} & $\mathbf{1.4\times}$ & - & $\mathbf{2.5\times}$ & - & $\mathbf{3.3\times}$ & - \\
   \hline
\end{tabular}
   }
\end{table}

\noindent For energy-efficiency we consider the operation of SNNAC in three feasible operating scenarios, {\em HighPerf} (high performance, maximum frequency), {\em EnOpt\_split} (energy optimal, disjoint logic and SRAM voltages), and {\em EnOpt\_joint} (energy optimal, unified voltage domains).
Figure \ref{fig:energy} shows the energy-per-cycle measurements on SNNAC for both logic and weight SRAMs, derived from test chip leakage and dynamic current measurements.
In {\em HighPerf}, operating frequency determines voltage settings, while frequency settings for {\em EnOpt\_split} and {\em EnOpt\_joint} are determined by the minimum-energy point (MEP) subject to voltage domain configurations.
The baselines for each operating scenario use the same clock frequencies and logic voltages as the optimized cases, but with SRAM operating at the nominal voltage.

In {\em HighPerf}, we observe that timing limitations prevent voltage scaling for the logic and memory in both the baseline and optimized configurations.
However, while the baseline is unable to scale SRAM voltage due to stability margins, the optimized case (with MATIC) is able to scale SRAM down to 0.65 V, resulting in 1.4$\times$ energy savings; timing requirements in the SRAM periphery prevent further scaling.

In {\em EnOpt\_split}, where SRAM and logic power rails are separated, the baseline is able to scale logic to the MEP but SRAM remains at the nominal voltage.
While the baseline is unable to voltage-scale its weight memories, with MATIC, we are able to scale both logic and SRAM to the MEP, leading to 2.5$\times$ energy savings.
Furthermore, SRAM energy is minimized at 0.5 V with a 28\% SRAM bit-cell failure rate, which corresponds to an 87\% classification rate on MNIST (versus 29.3\% for naive hardware).

Finally, in {\em EnOpt\_joint}, where voltage domains are unified to emulate a system with stringent power grid requirements, the baseline is unable to scale \textit{both} SRAM and logic voltages.
While logic voltage in the {\em HighPerf} scenario was limited due to timing requirements, logic in the baseline case for {\em EnOpt\_joint} is limited by SRAM $V_{min,read}$ since the power rails are shared;
in this case, \textit{SRAM PVT and read stability margins prevent system-wide voltage scaling.}
The MATIC-SNNAC combination, on the other hand, is able to scale both logic and SRAM voltages to the unified energy-optimal voltage, 0.55 V, which results in 3.3$\times$ energy savings.
The baseline design in {\em EnOpt\_split} is more efficient than the baseline design in {\em EnOpt\_joint}.
Although the relative savings-versus-baseline shows better results for {\em EnOpt\_joint}, the {\em EnOpt\_split} configuration provides the highest efficiency.
The energy-per-cycle measurements for the scenarios described above are summarized in Table~\ref{tab:energy}.

\begin{figure}[t]
\centering
\includegraphics[width=0.75\columnwidth]{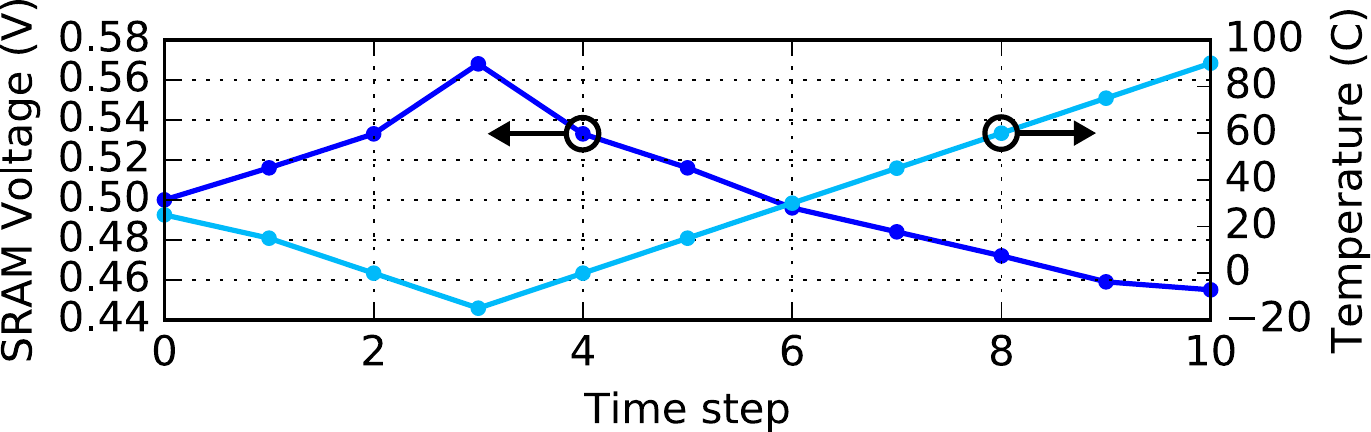}
\caption{
Runtime closed-loop SRAM voltage control enabled by the in-situ canary system, in response to ambient temperature variation.
The actual distance between time steps varies from 2-10 minutes due to the variable heating/cooling rate of the test chamber.
}
\label{fig:canary_temp}
\end{figure}

\begin{table}[t]
   \begin{adjustbox}{width=\columnwidth}
   \begin{threeparttable}
   \centering
   \caption{Comparison with State-Of-The-Art DNN Accelerators}
   \label{tab:perf}
   {\renewcommand{\arraystretch}{1.1}%
   \begin{tabular}{c"c|c|c"c|c}
      \cline{2-6}
      & \multicolumn{3}{c"}{ Low-power Fully-connected} & \multicolumn{2}{c}{ High-perf. Conv.} \\
      \cline{2-6}
                   & \textbf{This Work}$^*$   & ISSCC'17$^*$          & ISCA'16   & DATE'17         & ISSCC'16$^*$ \\
                   &                       & \cite{bang_dnn}   &\cite{eie} &\cite{chain_nn} &\cite{eyeriss} \\
      \thickhline
      Process    & \textbf{65 nm}         & 40 nm                     & 45 nm              & 28 nm                    & 65 nm \\
      \hline
      Area (mm sq.)  & \textbf{1.4}   & 7.1               & 0.64     & -              & 12.2 \\
      \hline
      DNN Type   & \textbf{Fully-conn.}           & Fully-conn. & Fully-conn. & Conv.                   & Conv. \\
      \hline
      Power (mW)  & \textbf{0.37}      & 0.29 & 9.2 & 33 & 567.5 \\
      \hline
      Frequency (MHz)  & \textbf{17.8}     & 3.9 & 800 & 204 & 700 \\
      \hline
      Voltage (V)    & \textbf{0.44-0.9}    & 0.63-0.9                & 1.0                & 0.9               & 0.82-1.17 \\
      \hline
      Energy (GOPS/W) & \textbf{119.2~/~400.5}$^\dagger$& 374                & 174         & 1421             & 243  \\
      \hline
   \end{tabular}
   }
   \begin{tablenotes}
      \item $^*$Performance metrics from a fabricated chip,
      \item $^\dagger$ Nominal energy efficiency / efficiency with MATIC
   \end{tablenotes}
   \end{threeparttable}
   \end{adjustbox}
   \end{table}

\subsection{Temperature Variation}
\noindent To demonstrate system stability over temperature, we execute the application benchmarks in a chamber with ambient temperature control, and sweep temperature from -15$^\circ$C to 90$^\circ$C for a given nominal voltage.
After initialization at the nominal voltage and temperature, we sweep the temperature down to -15$^\circ$C, and then up from -15$^\circ$C to 90$^\circ$C in steps of 15$^\circ$C, letting the chamber stabilize at each temperature point.
Figure \ref{fig:canary_temp} shows the SRAM voltage settings dictated by the in-situ canary system for an initial setting at 0.5 V on \textit{inversek2j}.
The results illustrate how the in-situ canary technique adjusts SRAM voltage to track temperature variation, while conventional systems would require static voltage margins.
We note that the operating voltages for the temperature chamber experiments are below the temperature inversion point for the 65 nm process; this is illustrated by the inverse relationship between temperature and SRAM voltage.

\subsection{Performance Comparison}
\noindent A comparison with recent DNN accelerator designs is listed in Table \ref{tab:perf}.
The performance comparison shows that the MATIC-SNNAC combination is comparable to state-of-the-art accelerators despite modest nominal performance, and enables a comparatively wider operating voltage range.
While the algorithmic characteristics (and hardware requirements) of networks containing convolutional layers are vastly different from FC-oriented DNNs~\cite{tpu_isca}, we include two recent Conv accelerators to show that SNNAC is competitive despite the lack of convolution-oriented optimization techniques.

\section{Related Work} \label{sec:related}
\noindent There has been a vast body of work on DNN hardware accelerators \cite{diannao, shidiannao, bang_dnn, chain_nn, eyeriss, eie, snnap} (see Section \ref{sec:intro}), as well as work addressing DNN fault tolerance.
Temam~\cite{temam_defect_tol} explores the impact of defects in logic gates, and is among the first to develop fault-tolerant hardware for neural networks.
Srinivasan et al.~\cite{mixedbits} exploit DNN resilience with a mixed 8T-6T SRAM where weight MSBs are stored in 8T cells.
However, this approach has no adaptation mechanism.
Xin et al.~\cite{xia_2017} and Liu et al.~\cite{liu_dac_17} design variability-tolerant training schemes, but for simulated resistive-RAM-based crossbars.
Yang and Murmann~\cite{yang_isqed} develop a noisy training scheme for Conv-DNNs, however noise is only added to input images.
In contrast, our work focuses on increasing the energy-efficiency of DNN accelerators in standard CMOS technologies by targeting a primary bottleneck in power dissipation, namely weight storage, and presents results from a fabricated DNN accelerator.

\section{Conclusions} \label{sec:conclusion}

\noindent This paper presents a methodology and algorithms that enable energy-efficient DNN accelerators to gracefully tolerate bit-errors from memory supply-voltage overscaling.
Our approach uses (1) memory-adaptive training - a technique that leverages the adaptability of neural networks to train around errors resulting from SRAM voltage scaling,
and (2) in-situ synaptic canaries - the use of bit-cells directly from weight SRAMs for voltage control and variation-tolerance.
To validate the effectiveness of MATIC, we designed and implemented SNNAC, a low-power DNN accelerator fabricated in 65 nm CMOS.
As demonstrated on SNNAC, the application of \codename{} results in either $3.3\times$ total energy reduction and $5.1\times$ energy reduction in SRAM, or $18.6\times$ reduction in application error.
Thus, \codename{} enables accurate inference on aggressively voltage-scaled DNN accelerators, and enables robust and efficient operation for a general class of DNN accelerators.

\section*{Acknowledgment}
\noindent\small{
   The authors would like to thank Fahim Rahman, Rajesh Pamula, and John Euhlin for design support.
   This work was supported in part by the National Science Foundation Grant CCF-1518703, generous gifts from Oracle Labs, Nvidia, and by C-FAR, one of the six SRC STARnet Centers, sponsored by MARCO and DARPA.
}

\bibliographystyle{IEEEtran}
\bibliography{paper}

\end{document}